\documentclass[10pt,twocolumn,letterpaper]{article}

\usepackage{cvpr}
\usepackage{times}
\usepackage{epsfig}
\usepackage{graphicx}
\usepackage{amsmath}
\usepackage{amssymb}
\usepackage{multirow}
\usepackage[usenames,dvipsnames]{color}
\usepackage{colortbl}
\definecolor{mygray}{gray}{.9}

\usepackage[pagebackref=true,breaklinks=true,letterpaper=true,colorlinks,bookmarks=false]{hyperref}

\cvprfinalcopy 

\def\cvprPaperID{****} 

\begin{document}

\title{Pose Invariant Embedding for Deep Person Re-identification}

\author{Liang Zheng$^{\dag}$, Yujia Huang$^{\ddag}$, Huchuan Lu$^{\S}$, Yi Yang$^{\dag}$ \\
 $^{\dag}$University of Technology Sydney \qquad $^{\ddag}$CMU   \qquad $^{\S} $Dalian University of Technology \qquad \\
{\tt\small {\{liangzheng06,yee.i.yang\}}@gmail.com yujiah1@andrew.cmu.edu lhchuan@dlut.edu.cn }
 }


\maketitle

\begin{abstract}
 Pedestrian misalignment, which mainly arises from detector errors and pose variations, is a critical problem for a robust person re-identification (re-ID) system. With bad alignment, the background noise will significantly compromise the feature learning and and matching process. To address this problem, this paper introduces the pose invariant embedding (PIE) as a pedestrian descriptor. First, in order to align pedestrians to a standard pose, the PoseBox structure is introduced, which is generated through pose estimation followed by affine transformations. Second, to reduce the impact of pose estimation errors and information loss during PoseBox construction, we design a PoseBox fusion (PBF) CNN architecture that takes the original image, the PoseBox, and the pose estimation confidence as input. The proposed PIE descriptor is thus defined as the fully connected layer of the PBF network for the retrieval task. Experiments are conducted on the Market-1501, CUHK03, and VIPeR datasets. We show that PoseBox alone yields decent re-ID accuracy, and that when integrated in the PBF network, the learned PIE descriptor produces competitive performance compared with the state-of-the-art approaches.
\end{abstract}

\section{Introduction}\label{sec:intro}
This paper studies the task of person re-identification (re-ID). Given a probe (person of interest) and a gallery, we aim to find  in the gallery all the images containing the same person with the probe. We focus on the identification problem, a retrieval task in which each probe has at least one ground truth in the gallery \cite{zheng2015scalable}. A number of factors affect the re-ID accuracy, such as detection/tracking errors, variations in illumination, pose, viewpoint, \emph{etc}.

\begin{figure}
  \centering
  \includegraphics[width=3.2in]{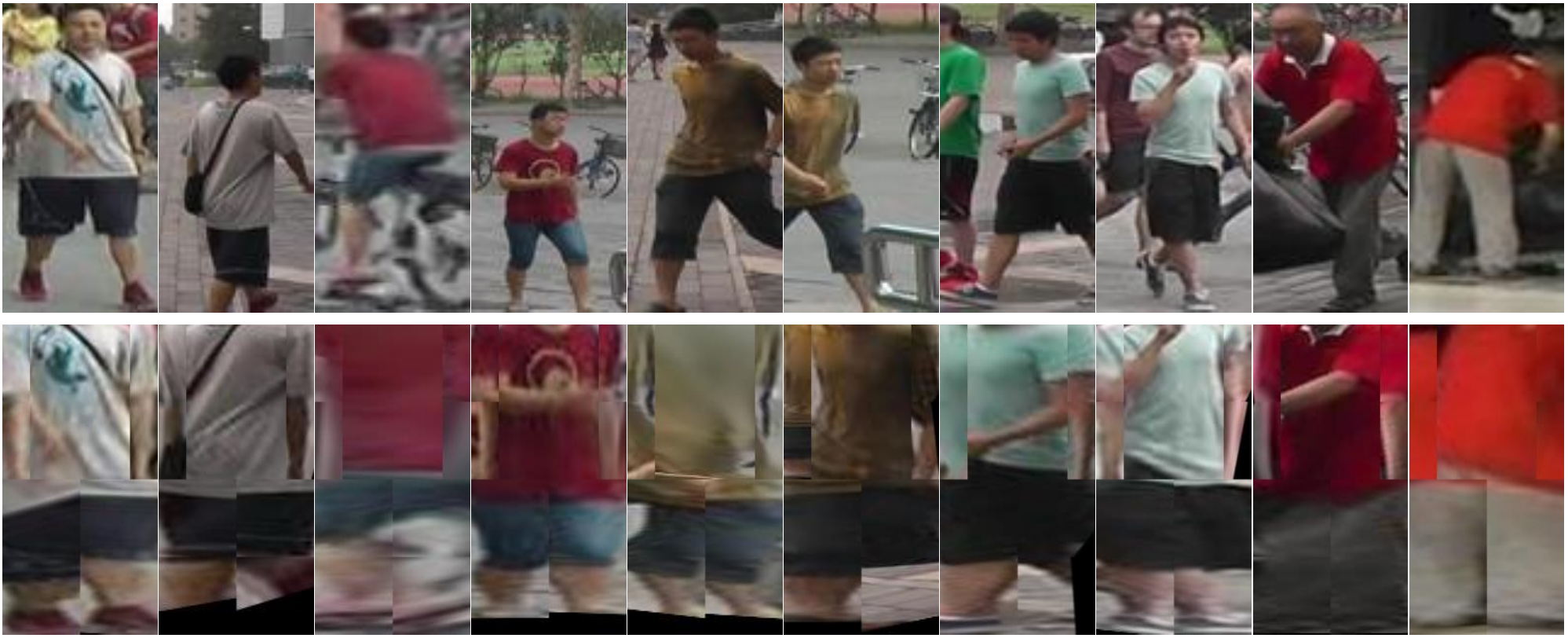}\\
  \caption{Examples of misalignment correction by PoseBox. Row 1: original bounding boxes with detection errors/occlusions. Every consecutive two boxes correspond to a same person. Row 2: corresponding PoseBoxes. We observe that misalignment can be corrected to some extent. }\label{fig:corrected_misalignment}
\end{figure}

A critical influencing factor on re-ID accuracy is the misalignment of pedestrians, which can be attributed to two causes. First, pedestrians naturally take on various poses as shown in Fig. \ref{fig:corrected_misalignment}. Pose variations imply that the position of the body parts within the bounding box is not predictable. For example, it is possible that one's hands reach above the head, or that one is riding a bicycle instead of being upright. The second cause of misalignment is detection error. As illustrated in the second row of Fig. \ref{fig:corrected_misalignment}, detection errors may lead to severe vertical misalignment.

When pedestrians are poorly aligned, the re-ID accuracy can be compromised. For example, a common practise in re-ID is to partition the bounding box into horizontal stripes \cite{li2014deepreid,zheng2015scalable,ahmed2015improved,liao2015person}. This method works under the assumption of slight vertical misalignment. But when vertical misalignment does happen as in the cases in Row 2 of Fig. \ref{fig:corrected_misalignment}, one's head will be matched to the background of a misaligned image. So horizontal stripes may be less effective when severe misalignment happens. In another example, under various pedestrian poses, the background may be incorrectly weighted by the feature extractors and thus affect the following matching accuracy.

To our knowledge, two previous works \cite{Cheng2011custom,cheng2014person} from the same group explicitly consider the misalignment problem. In both works, the pictorial structure (PS) is used, which shares a similar motivation and construction process with PoseBox, and the retrieval process mainly relies on matching the normalized body parts. While the idea of constructing normalized poses is similar, our work locates body joints using a state-of-the-art CNN based pose estimator, and the components of PoseBox are different from PS as evidenced by large-scale evaluations. Another difference of our work is the matching procedure. While \cite{Cheng2011custom,cheng2014person} do not discuss the pose estimation errors which prevalently exist in real-world datasets, we show that these errors make rigid feature learning/matching with only the PoseBox yield inferior results to the original image, and that the three-stream PoseBox fusion network effectively alleviates this problem.


Considering the above-mentioned problems and the limit of previous methods, this paper proposes the pose invariant embedding (PIE) as a robust visual descriptor. Two steps are involved. First, we construct a PoseBox for each pedestrian bounding box. PoseBox depicts a pedestrian with standarized upright stance. Carefully designed with the help of pose estimators \cite{wei2016convolutional}, PoseBox aims to produce well-aligned pedestrian images so that the learned feature can find the same person under intensive pose changes. Trained alone using a standard CNN architecture \cite{xiao2016learning,zheng2016mars,zheng2016prw}, we show that PoseBox yields very decent re-ID accuracy.

\begin{figure}
  \centering
  \includegraphics[width=3.2in]{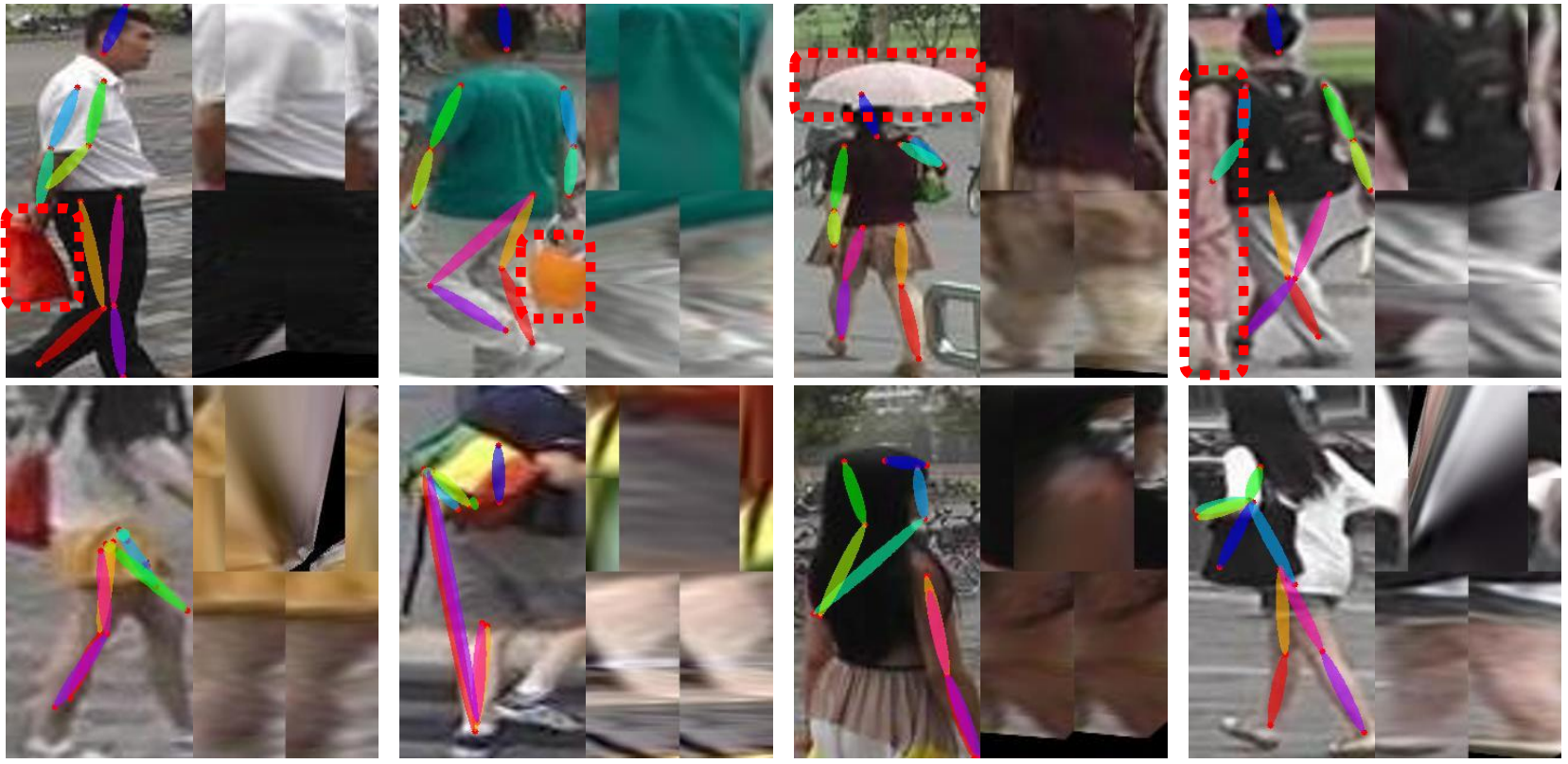}\\
  \caption{Information loss and pose estimation errors that occur during PoseBox construction. Row 1: important pedestrian details (highlighted in red bounding boxes) may be missing in the PoseBox. Row 2: pose estimation errors deteriorate the quality of PoseBoxes. For each image pair, the original image and its PoseBox are on the left and right, respectively. }\label{fig:errorANDmiss}
\end{figure}

Second, to reduce the impact of information loss and pose estimation errors (Fig. \ref{fig:errorANDmiss}) during PoseBox construction, we build a PoseBox fusion (PBF) CNN model with three streams as input: the PoseBox, the original image, and the pose estimation confidence. PBF achieves a globally optimized tradeoff between the original image and the PoseBox. PIE is thus defined as the FC activations of the PBF network. On several benchmark datasets, we show that the joint training procedure yields competitive re-ID accuracy to the state of the art. To summarize, this paper has three contributions.

\vspace{-0.2cm}
\begin{itemize}
  \item Minor contribution: the PoseBox is proposed which shares a similar nature with a previous work \cite{Cheng2011custom}. It enables well-aligned pedestrian matching, and yields satisfying re-ID performance when being used alone.
\vspace{-0.2cm}
  \item Major contribution: the pose invariant embedding (PIE) is proposed as a part of the PoseBox Fusion (PBF) network. PBF fuses the original image, PoseBox and the pose estimation errors, thus providing a fallback mechanism when pose estimation fails.
\vspace{-0.2cm}
  \item Using PIE, we report competitive re-ID accuracy on the Market-1501, CUHK03, and VIPeR datasets.
\end{itemize}

\section{Related Work} \label{sec:related_work}

\textbf{Pose estimation.} The pose estimation research has shifted from traditional approaches \cite{Cheng2011custom,cheng2014person} to deep learning following the pioneer work ``DeepPose'' \cite{toshev2014deeppose}. Some recent methods employ multi-scale features and study mechanisms on how to combine them \cite{tompson2014joint,newell2016stacked}. It is also effective to inject spatial relationships between body joints by regularizing the unary scores and pairwise comparisons \cite{fan2015combining,pishchulin2015deepcut}. This paper adopts the convolutional pose machines (CPM) \cite{wei2016convolutional}, a state-of-the-art pose estimator with multiple stages and successive pose predictions.

\textbf{Deep learning for re-ID.} Due to its superior performance, deep learning based methods have been dominating the re-ID community in the past two years. In the two earlier works \cite{li2014deepreid,yi2014deep}, the siamese model which takes two images as input is used. In later works, this model is improved in various ways, such as injecting more sophisticated spatial constraint \cite{ahmed2015improved,cheng2016person}, modeling the sequential properties of body parts using LSTM \cite{varior2016siamese}, and mining discriminative matching parts for different image pairs \cite{varior2016gated}. It is pointed out in \cite{zheng2016survey} that the siamese model only uses weak re-ID labels: two images being of the same person or not; and it is suggested that an identification model which fully uses the strong re-ID labels be superior. Several previous works adopt the identification model \cite{xiao2016learning,wu2016enhanced,zheng2016mars}. In \cite{zheng2016mars}, the video frames are used as training samples of each person class, and in \cite{xiao2016learning}, effective neurons are discovered for each training domain and a new dropout strategy is proposed. The architecture proposed in \cite{wu2016enhanced} is more similar to the PBF model in our work. In \cite{wu2016enhanced}, hand-crafted low-level features are concatenated after a fully connected (FC) layer which is connected to the softmax layer. Our network is similar to \cite{wu2016enhanced} in that confidence scores of pose estimation are catenated with the other two FC layers. It departs from \cite{wu2016enhanced} in that our network takes three streams as input, two of which are raw images.

\textbf{Poses for re-ID.} Although pose changes have been mentioned by many previous works as an influencing factor on re-ID, only a handful of reports can be found discussing the connection between them. Farenzena \emph{et al.} \cite{farenzena2010person} propose to detect the symmetrical axis of different body parts and extract features following the pose variation. In \cite{weinrich2013appearance}, rough estimates of the upper-body orientation is provided by the HOG detector, and the upper body is then rendered into the texture of an articulated 3D model. Bak \emph{et al.}  \cite{bak2015person} further classify each person into three pose types: front, back, and side. A similar idea is exploited in \cite{cho2016improving}, where four pose types are used. Both works \cite{bak2015person,cho2016improving} apply view-point specific distance metrics according to different testing pose pairs. The closest works to PoseBox are \cite{Cheng2011custom,cheng2014person}, which construct the pictorial structure (PS), a similar concept to PoseBox. They use traditional pose estimators and hand-crafted descriptors that are inferior to CNN by a large margin. Our work employs a full set of stronger techniques, and designs a more effective CNN structure evidenced by the competitive re-ID accuracy on large-scale datasets.

\section{Proposed Method}\label{sec:method}

\begin{figure}
  \centering
  \includegraphics[width=2.7in]{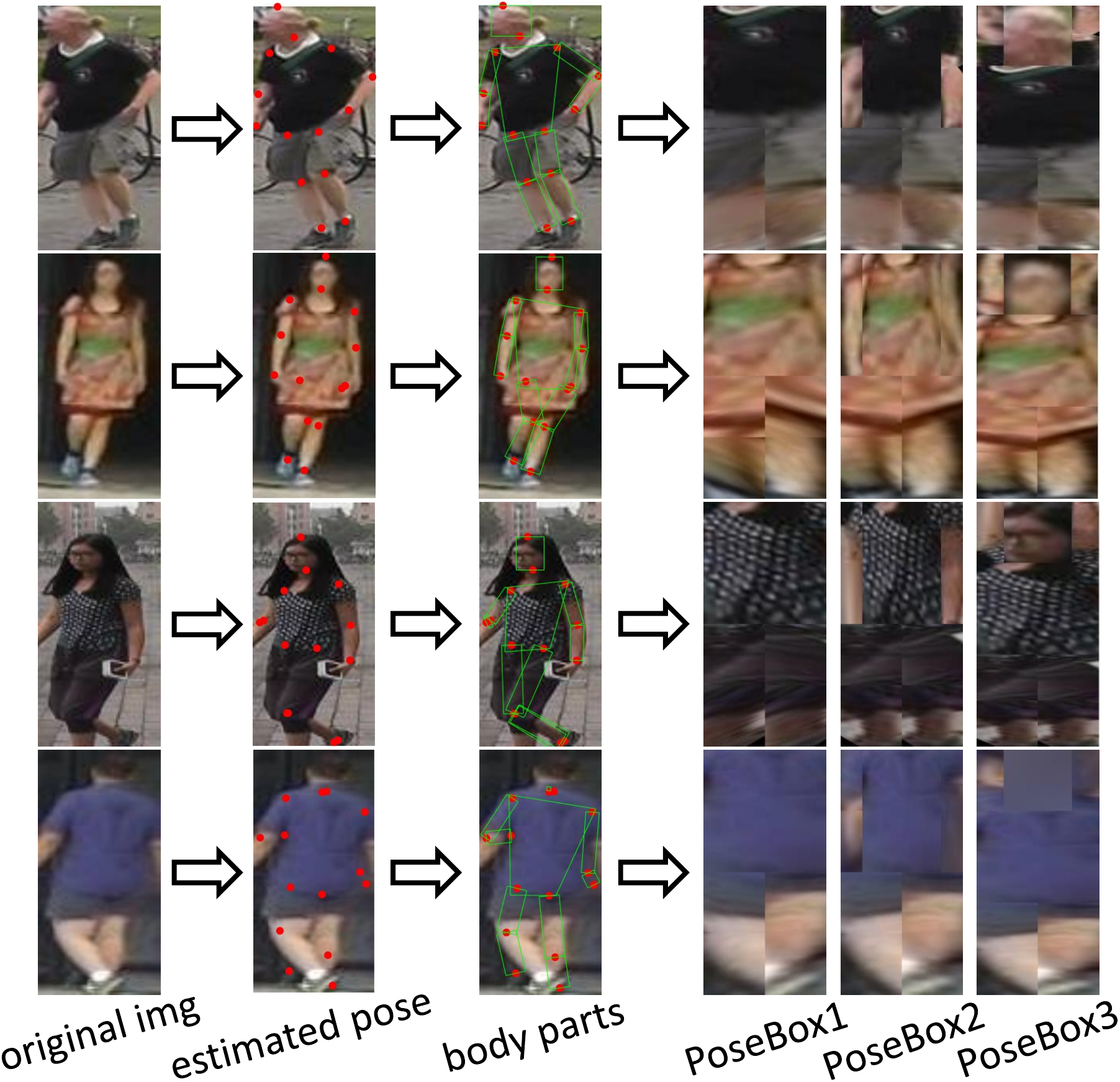}\\
  \caption{PoseBox construction. Given an input image, the pedestrian pose is estimated by CPM \cite{wei2016convolutional}. Ten body parts can then be discovered through the body joints. Three types of PoseBoxes are built from the body parts. PoseBox1: torso + legs; PoseBox2: PoseBox1 + arms; PoseBox3: PoseBox2 + head. }\label{fig:pose_construction}
\end{figure}
\subsection{PoseBox Construction}\label{sec:posebox}
The construction of PoseBox has two steps, \emph{i.e.,} pose estimation and PoseBox projection.

\textbf{Pose estimation.} This paper adopts the off-the-shelf model of the convolutional pose machines (CPM) \cite{wei2016convolutional}. In a nutshell, CPM is a sequential convolutional architecture that enforces intermediate supervision to prevent vanishing gradients. A set of 14 body joints are detected, \ie, head, neck, left and right shoulders, left and right elbows, left and right wrists, left and right hips, left and right knees, and left and right ankles, as shown in the second column of Fig. \ref{fig:pose_construction}

\textbf{Body part discovery and affine projection.} From the detected joints, 10 body parts can be depicted (the third column of Fig. \ref{fig:pose_construction}). The parts include head, torso, upper and lower arms (left and right), and upper and lower legs (left and right), which almost cover the whole body. These quadrilateral parts are projected to rectangles using affine transformations.

In more details, the head is defined with the joints of head and neck, and we manually set the width of each head box to $\frac{2}{3}$ of its height (from head to neck). An upper arm is confined by the shoulder and elbow joints, and the lower arm by the elbow and wrist joints. The width of the arms boxes is set to 20 pixels. Similarly, the upper and lower legs are defined by the hip and knee joints, and the knee and ankle joints, respectively. Their widths are both 30 pixels. The torso is confined by four body joints, \ie, the two shoulders and the two hips, so we simply draw a quadrangle for the torso. Due to pose estimation errors, the affine transformation may encounter singular values. So in practice, we add some small random disturbance when the pose estimation confidence of a body part is below a threshold (set to 0.4).

\textbf{Three types of PoseBoxes.} In several previous works discussing the performance of different parts, a common observation is that the torso and legs make the largest contributions \cite{Cheng2011custom,ahmed2015improved,cheng2016person}. This is expected because the most distinguishing features exist in the upper-body and lower-body clothes. Based on the existing observations, this paper builds three types of PoseBoxes as described below.
\vspace{-0.2cm}

\begin{itemize}
  \item PoseBox 1. It consists of the torso and two legs. A leg is comprised of the upper and the lower legs. PoseBox 1 includes two most important body parts  and is a baseline for the other two PoseBox types.
    \vspace{-0.2cm}
  \item PoseBox 2. Based on PoseBox 1, we further add the left and right arms. An arm includes the upper and lower arm sub-modules. In our experiment we show that PoseBox 2 is superior to PoseBox 1 due to the enriched information brought by the arms.
    \vspace{-0.2cm}
  \item PoseBox 3. On the basis of PoseBox 2, we put the head box on top of the torso box. It is shown in \cite{Cheng2011custom} that the inclusion of head brought marginal performance increase. In our case, we find that PoseBox 3 slightly inferior to PoseBox2, probably because of the frequent head/neck estimation errors.
\end{itemize}

\textbf{Remarks.} The advantage of PoseBox is two-fold. First, the pose variations can be corrected. Second, background noise can be removed largely.

PoseBox is also limited in two aspects. First, pose estimation errors often happen, leading to imprecisely detected joints. Second, PoseBox is designed manually, so it is not guaranteed to be optimal in terms of information loss or re-ID accuracy. We address the two problems by a fusion scheme to be introduced in Section \ref{sec:joint_training}. For the second problem, specifically, we note that we construct PoseBoxes \emph{manually} because current re-ID datasets do not provide ground truth poses, without which it is not trivial to design an end-to-end learning method to automatically generate normalized poses.

\subsection{Baselines}\label{sec:baseline}
\begin{figure}
  \centering
  \includegraphics[width=2.6in]{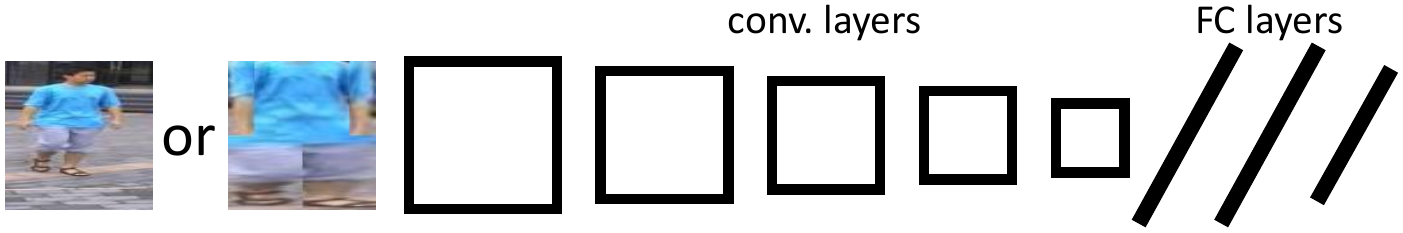}\\
  \caption{The baseline identification CNN model used in this paper. The AlexNet \cite{krizhevsky2012imagenet} or ResNet-50 \cite{he2016deep} with softmax loss is used. The FC activations are extracted for Euclidean-distance testing.}\label{fig:baseline}
\end{figure}

This paper constructs two baselines based on the original pedestrian image and PoseBox, respectively. According to the results in the recent survey \cite{zheng2016survey}, the identification model \cite{krizhevsky2012imagenet} outperforms the verification model \cite{ahmed2015improved,li2014deepreid} significantly on the Market-1501 dataset \cite{zheng2015scalable}: the former makes full use of the re-ID labels, \ie, the identity of each bounding box, while the latter only uses weak labels, \ie, whether two boxes belong to the same person. So in this paper we adopt the identification CNN model (Fig. \ref{fig:baseline}). Specifically, this paper uses the standard AlexNet \cite{krizhevsky2012imagenet} and Residual-50 \cite{he2016deep} architectures. We refer readers to the respective papers for detailed network descriptions.

During training, we employ the default parameter settings, except editing the last FC layer to have the same number of neurons as the number of distinct IDs in the training set. During testing, given an input image resized to $224\times224$, we extract the FC7$/$FC8 activations for AlexNet, and the Pool5$/$FC activations for ResNet-50. After $\ell_2$ normalization, we use Euclidean distance to perform person retrieval in the testing set. With respect to the input image type, two baselines are used in this paper:
\vspace{-0.2cm}
\begin{itemize}
  \item Baseline1: the original image (resized to $224\times224$) is used as input to CNN during training and testing.
      \vspace{-0.2cm}
  \item Baseline2: the PoseBox (resized to $224\times224$) is used as input to CNN during training and testing. Note that only one PoseBox type is used each time.
\end{itemize}
\vspace{-0.2cm}

\subsection{The PoseBox Fusion (PBF) Network}\label{sec:joint_training}
\begin{figure}
  \centering
  \includegraphics[width=3.2in]{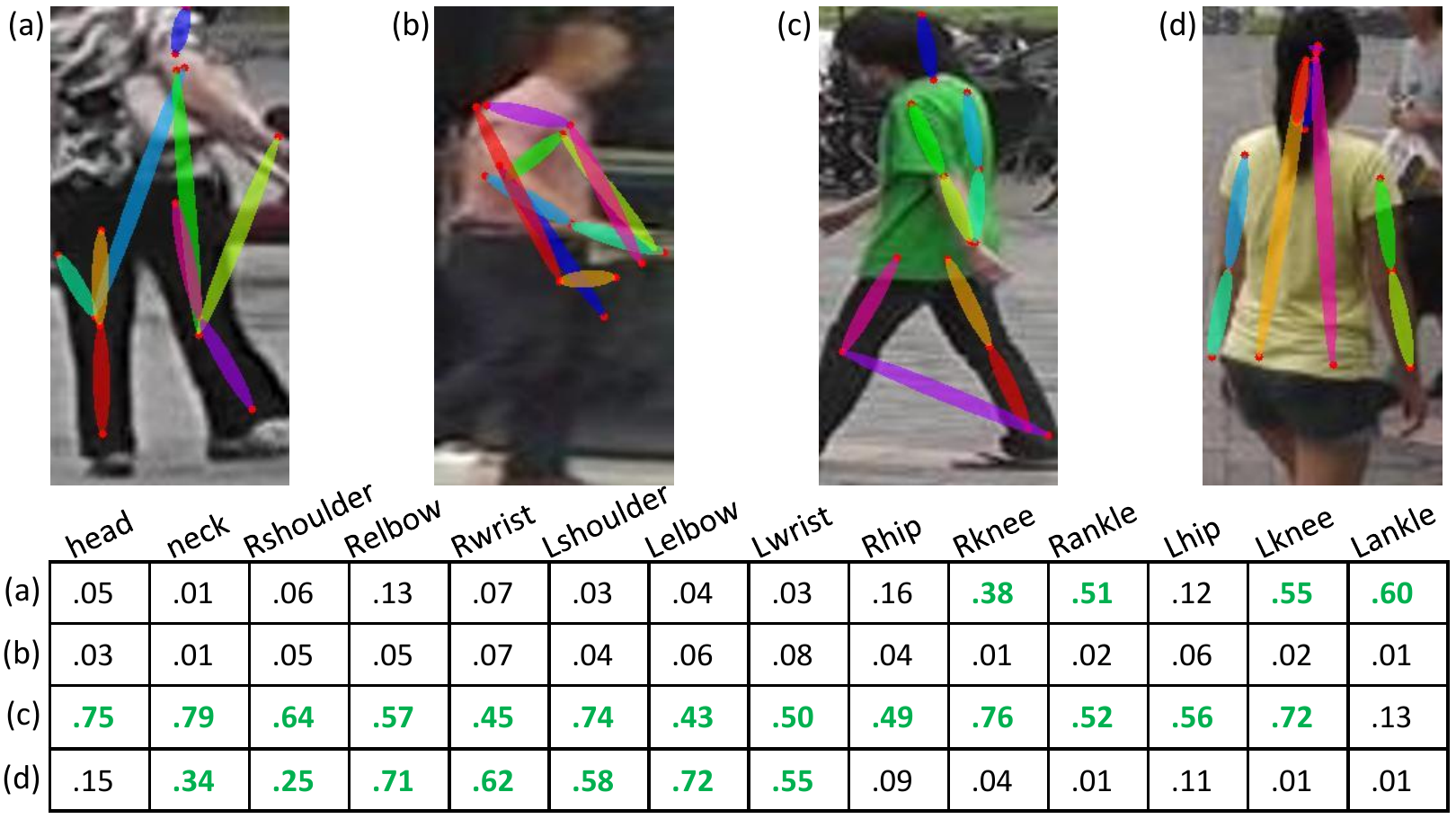}\\
  \caption{Examples of pose estimation errors and the confidence scores. \textbf{Upper:} four pedestrian bounding boxes named with (a), (b), (c), and (d),  and their pose estimation results. \textbf{Lower:} pose estimation confidence scores of the four images. A confidence vector consists of 14 numbers corresponding to the 14 body joints. We highlight the correctly detected joints in \textcolor{green}{green}. }\label{fig:pose_wrong}
\end{figure}

\textbf{Motivation.}
During PoseBox construction, pose estimation errors and information loss may happen, leading to compromised quality of the PoseBox (see Fig. \ref{fig:errorANDmiss}). On the one hand, pose estimation errors often happen, as we use an off-the-shelf pose estimator (which is usually the case under practical usage). As illustrated in Fig. \ref{fig:pose_wrong} and Fig. \ref{fig:corrected_misalignment}, pose estimation may fail when the detections have missing parts or the pedestrian images are of low resolution. On the other hand, when cropping human parts from a bounding box, it is inevitable that important details are missed out, such as bags and umbrellas (Fig. \ref{fig:errorANDmiss}). The failure in the construction of high-quality PoseBoxes and the information loss during part cropping may result in compromised results of the baseline 2. This is confirmed in the experiment that baseline 1 yields superior re-ID accuracy to baseline 2.

For the first problem, \ie, the pose estimation errors, we can mostly foretell the quality of pose estimation by resorting to the confidence scores (examples can be seen in Fig. \ref{fig:pose_wrong}). Under high estimation confidence, we envision fine quality of the generated PoseBox. But when the pose estimation confidence scores are low for some body parts, it may be expected that the constructed PoseBox has poor quality. For the second problem, the missing visual cues can be rescued by re-introducing the original image, so that the discriminative details are captured by the deep network.

\begin{figure*}[t]
  \centering
  \includegraphics[width=6.2in]{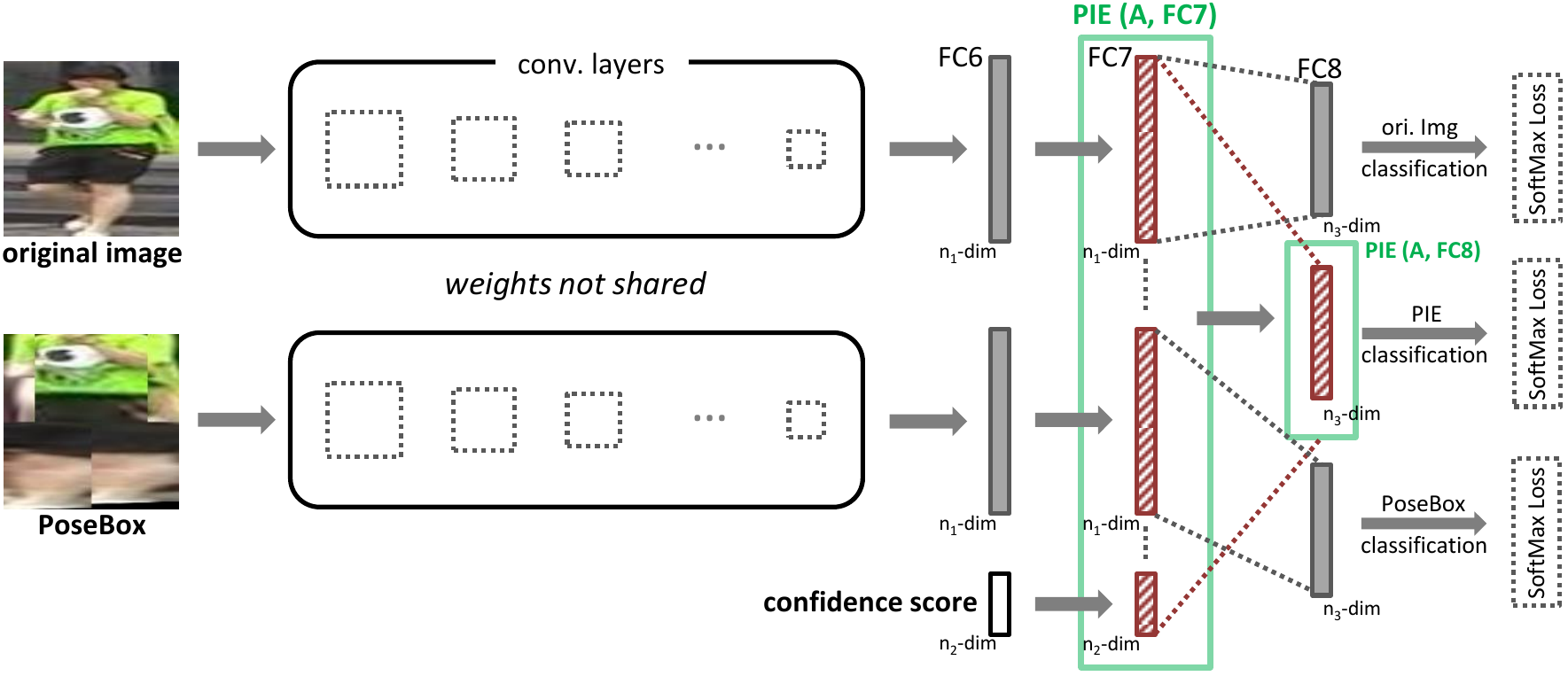}\\
  \caption{Illustration of the PoseBox Fusion (PBF) network using AlexNet. The network inputs, \ie, the original image, its PoseBox, and the pose estimation confidence, are highlighted in \textbf{bold}. The former two undergo convolutional layers before the fully connections (FC). The confidence vector undergoes one FC layer, before the three FC7 layers are concatenated and fully connected to FC8. SoftMax loss is used. Two alternatives of the PIE descriptor are highlighted by green boxes. For AlexNet and Market-1501, PIE(A, FC7) is 8,206-dim, and PIE(A, FC8) is 751-dim; For ResNet-50, there would be no FC6, and PIE(R, Pool5) is 4,110-dim, and PIE(R, FC) is 751-dim. }\label{fig:fusion_network}
\end{figure*}

\textbf{Network.} Given the above considerations, this paper proposes a three-stream PoseBox Fusion (PBF) network which takes the original image, the PoseBox, and the confidence vector as input (see Fig. \ref{fig:fusion_network}). To leverage the ImageNet \cite{deng2009imagenet} pre-trained models, two types of image inputs, \ie, the original image and the PoseBox are resized to $256\times256$ (then cropped randomly to $227\times227$) for AlexNet \cite{krizhevsky2012imagenet} and $224\times224$ for ResNet-50 \cite{he2016deep}. The third input, \ie, pose estimation confidence scores, is a 14-dim vector, in which each entry falls within the range $[0, 1]$.

The two image inputs are fed to two CNNs of the same structure. Due to the content differences of the original image and its PoseBox, the two streams of convolutional layers do not share weights, although they are initialized from the same seed model. The FC6 and FC7 layers are connected to these convolutional layers. For the confidence vector, we add a small FC layer which projects the 14-dim vector to a 14-dim FC vector. We concatenate the three inputs at the FC7 layer, which is further fully connected to FC8. The sum of the three Softmax losses is used for loss computation. When the ResNet-50 \cite{he2016deep} is used instead of AlexNet, Fig. \ref{fig:fusion_network} does not have the FC6 layers, and the FC7 and FC8 layers are known as Pool5 and FC.

In Fig. \ref{fig:fusion_network}, as denoted by the green bounding box, the pose invariant embedding (PIE) can either be the concatenated FC7 activations (4,096+4,096+14 = 8,206-dim) or its next fully connected layer (751-dim and 1,160-dim for Market-1501 and CUHK03, respectively). For AlexNet, we denote the two PIE descriptors as PIE(A, FC7) and PIE(A, FC8), respectively; for ResNet-50, they are termed as PIE(R, Pool5) and PIE(R, FC), respectively.

During training, batches of the input triplets (the original image, its PoseBox, and the confidence vector) are fed into PBF, and the sum of the three losses is back-propagated to the convolutional layers. The ImageNet pretrained model initializes both the original image and PoseBox streams.

During testing, given the three inputs of an image, we extract PIE as the descriptor. Note that we apply ReLU on the extracted embeddings, which produces superior results according to our preliminary experiment. Then the Euclidean distance is used to calculate the similarity between the probe and gallery images, before a sorted rank list is produced.

PBF has three advantages. First, the confidence vector is an indicator whether PoseBox is reliable. This improves the learning ability of PBF as a static embedding network, so that a global tradeoff between the PoseBox and the original image can be found.
Second, the original image not only enables a fallback mechanism when pose estimation fails, but also retrains the pedestrian details that may be lost during PoseBox construction but are useful in discriminating identities. Third, the PoseBox provides important complementary cues to the original image. Using the correctly predicted joints, pedestrian matching can be more accurate with the well-aligned images. The influence of detection errors and pose variations can thus be reduced.

\section{Experiment}\label{sec:experiment}
\subsection{Dataset}
This paper uses three datasets for evaluation, \ie, VIPeR \cite{gray2007evaluating}, CUHK03 \cite{li2014deepreid}, and Market-1501 \cite{zheng2015scalable}. The \textbf{VIPeR} dataset contains 632 identities, each having 2 images captured by 2 cameras. It is evenly divided into training and testing sets, each consisting of 316 IDs and 632 images. We perform 10 random train/test splits and calculate the averaged accuracy. The \textbf{CUHK03} dataset contains 1,360 identities and 13,164 images. Each person is observed by 2 cameras, and on average there are 4.8 images under each camera. We adopt the single-shot mode and evaluate this dataset under 20 random train/test splits. The \textbf{Market-1501} dataset is featured by 1,501 IDs, 19,732 gallery images and 12,936 training images captured by 6 cameras. Both CUHK03 and Market-1501 are produced by the DPM detector \cite{fernando2012discriminative}. The Cumulative Matching Characteristics (CMC) curve is used for all the three datasets, which encodes the possibility that the query person is found within the top $n$ ranks in the rank list. For Market-1501 and CUHK03, we additionally employ the mean Average Precision (mAP), which considers both the precision and recall of the retrieval process \cite{zheng2015scalable}. The evaluation toolbox provided by the Market-1501 authors is used.

\begin{table*}[!t]
\footnotesize
\setlength{\tabcolsep}{5.8pt}
\caption{Comparison of the proposed method with various baselines. PoseBox2 is employed here. Baseline1: training using the original image. Baselin2: training use the PoseBox. PIE: proposed pose invariant embedding. A: AlexNet. R: ResNet-50.}
\begin{center}
\begin{tabular}{l|l|ccccc|cccc|cccc}\hline
\multirow{2}{*}{Methods}& \multirow{2}{*}{dim} & \multicolumn{5}{c|}{\emph{Market-1501}}   & \multicolumn{4}{c|}{\emph{CUHK03}} &
\multicolumn{4}{c}{\emph{Market-1501 $\to$ VIPeR}} \\\cline{3-15}
& & 1 &5&10&20&mAP&1&5&10&20&1&5&10&20\\
\hline
\hline
 Baseline1 (A, FC7) &4,096 & 55.49&76.28 &83.55&88.98 & 32.36& 57.15&83.50 &90.85&95.70&17.44&31.84&41.04& 51.36  \\
 Baseline1 (A, FC8) &751 &53.65 & 75.48&82.93 &88.51& 31.60& 58.80&85.80 &91.90&96.25&17.15&32.06&41.68& 51.55    \\
 Baseline1 (R, Pool5) &2,048 & 73.02 & 87.44  &91.24 &94.70& 47.62&51.60&79.60 &87.70&95.00 & 23.42&42.31&51.96&63.80   \\
 Baseline1 (R, FC) &751 &70.58 & 84.95 & 90.02&93.53&45.84&54.80&84.20 &91.70&97.60&15.85&28.80&37.41&47.85 \\

\hline
Baseline2 (A, FC7) & 4,096& 52.22& 71.53&78.95 &85.04&28.95 & 39.90&71.40&82.30&90.00&17.28&32.59&42.25& 55.09    \\
Baseline2 (A, FC8) & 751& 51.10&72.24 &79.48&85.60 & 29.91& 42.30& 75.05 &84.35&92.00&16.04&33.45&42.66& 54.97   \\
Baseline2 (R, Pool5) & 2,048& 64.49&79.48 &85.07&88.95 &38.16 &36.90 &68.40 &78.70&86.70& 21.11&37.18&45.89&54.34  \\
Baseline2 (R, FC) & 751&62.20 & 78.36&83.76& 88.84&37.91 &41.70 &72.70 &84.20&92.50&15.57&26.68&33.54&41.71     \\
\hline
\hline
PIE (A, FC7) & 8,206&64.61 & 82.07&87.83&91.75 &38.95 &59.80&85.35 &91.85&95.85&21.77&38.04&46.61& 56.61    \\
PIE (A, FC8) & 751&65.68 & 82.51&87.89&91.63 &41.12 &\textbf{62.40}&88.00 &93.70&96.50&18.10&31.20&38.92& 49.40  \\
PIE (R, Pool5) & 4,108& \textbf{78.65} & \textbf{90.26}&\textbf{93.59}& \textbf{95.69}& \textbf{53.87}&57.10 &84.60 &91.40&96.20& \textbf{27.44}&\textbf{43.01}&\textbf{50.82}&\textbf{60.22}     \\
PIE (R, FC) & 751&75.12 &88.27 &92.28&94.77 &51.57 &61.50 &\textbf{89.30} &\textbf{94.50}&\textbf{97.60}& 23.80&37.88&47.31&56.55   \\

\hline
\end{tabular}
\end{center}
\label{table:various_approaches}
\end{table*}
\subsection{Experimental Setups}
Our experiments directly employ the off-the-shelf convolutional pose machines (CPM) trained using the multi-stage CNN model trained on the MPII human pose dataset \cite{andriluka20142d}. Default settings are used with input images resized to $384\times192$. For the PBF network, we replace the convolutional layers with those from either the AlexNet \cite{krizhevsky2012imagenet} or ResNet-50 \cite{he2016deep}. When AlexNet is used, $n_1 = 4,096, n_2 = 14, n_3 = 751$. When ResNet-50 is used, PBF will not have the FC6 layer, and the FC7 layer is denoted by Pool5: $n_1 = 2,048, n_3 = 751$. We train the PBF network for 36 epochs. The initial learning rate is set to 0.01, and is reduced by 10x every 6 epochs. We run the deep learning experiments using GTX1080 under the Caffe framework \cite{jia2014caffe} and the batch size is set to 32 and 16 using AlexNet and ResNet-50, respectively. For both CNN models, it takes 6-7 hours for the training process to converge on the Market-1501 dataset.

We train PIE on Market-1501 and CUHK03, respectively, which have relatively large data volumes. We also test the generalization ability of PIE on some smaller datasets such as VIPeR. That is, we only extract features using the model pre-trained on Market-1501, and then learn some distance metric on the small datasets.

\subsection{Evaluation}
\textbf{Baselines.}
We first evaluate the the two re-ID baselines described in Section \ref{sec:baseline}. The results on three datasets are shown in Table \ref{table:various_approaches}. Two major conclusions can be drawn.

First, we observe that very competitive performance can be achieved by baseline 1, \ie, training with the original image. Specifically, on Market-1501, we achieve rank-1 accuracy of 55.49\% and 73.02\% using AlexNet and ResNet-50, respectively. These numbers are consistent with those reported in \cite{zheng2016survey}. Moreover, we find that FC7 (Pool5) is superior to FC8 (FC) on Market-1501 but situation reverses on CUHK03. We speculate the CNN model is trained to be more specific to the Market-1501 training set due to its larger data volume, so retrieval on Market-1501 is more of a transfer task than CUHK03. This is also observed in transferring ImageNet models to other recognition tasks \cite{sharif2014cnn}.

\begin{figure}
  \centering
  \includegraphics[width=2.9in]{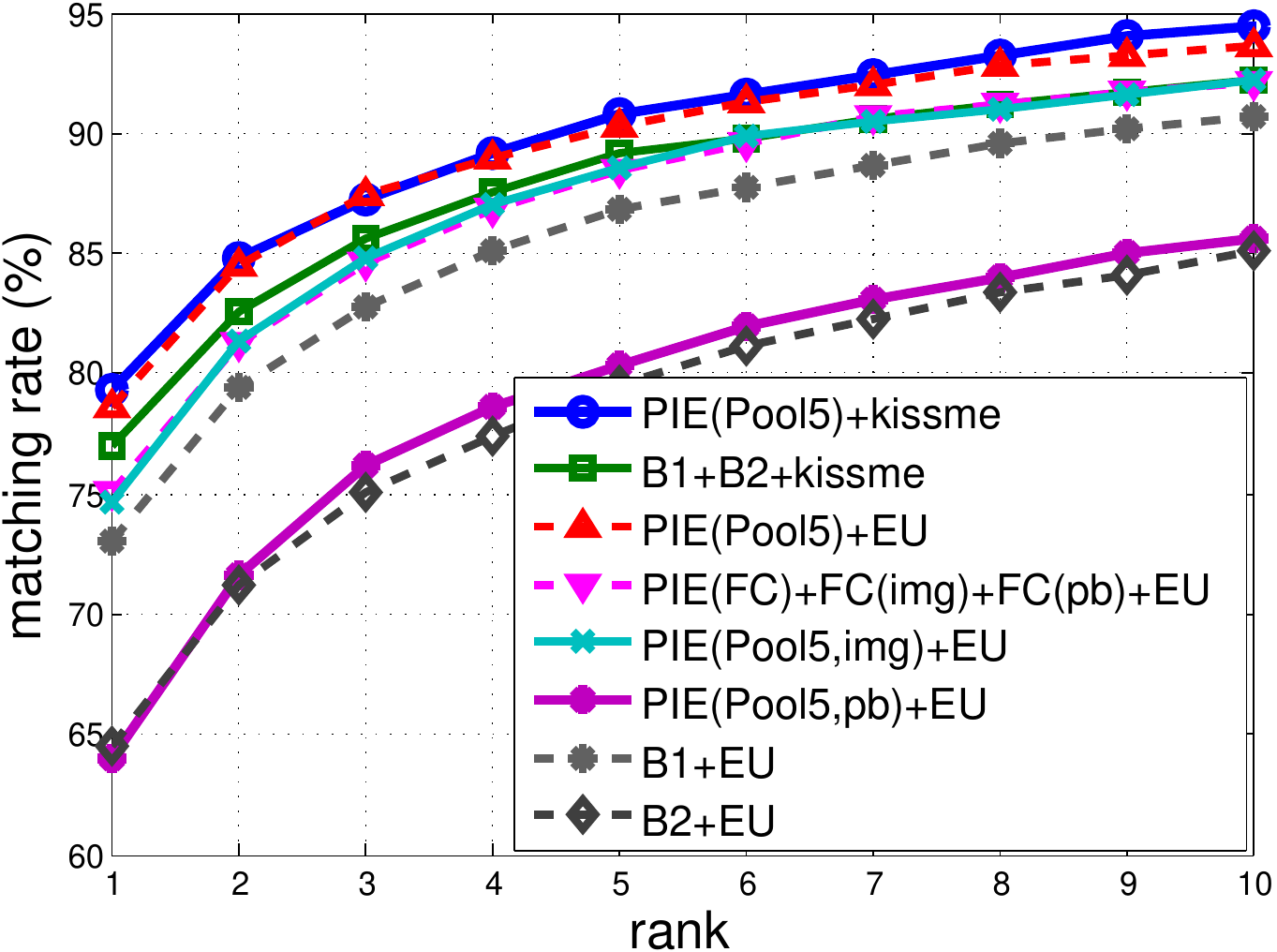}\\
  \caption{Comparison with various feature combinations on the Market-1501 dataset. ResNet-50 \cite{he2016deep} is used. Kissme \cite{kostinger2012large} is used for distance metric learning. ``EU'': Euclidean distance. ``PIE(Pool5,img)'' and ``PIE(Pool5,pb)'' denote the 2,048-dim sub-vectors of the full 4,108-dim PIE(Pool5) vector, corresponding to the image and PoseBox streams of PBF, respectively. ``FC(img)'' and ``FC(pb)'' are the 751-dim FC vectors of the image and PoseBox streams of PBF, respectively. ``B1'' and ``B2'' represent baseline 1 and 2, respectively, using the 2,048-dim Pool5 features.}\label{fig:comparisons}
\end{figure}
Second, compared with baseline 1, we can see that baseline 2 is to some extent inferior. On the Market-1501 dataset, for example, results obtained by baseline 2 is 3.3\% and 8.9\% lower using AlexNet and ResNet-50, respectively. The performance drop is expected due to the pose estimation errors and information loss mentioned in Section \ref{sec:joint_training}. Since this paper only employs the off-the-shelf pose estimator, we speculate in the future that the PoseBox baseline can be improved by re-training pose estimation using newly labeled data on the re-ID datasets.

\textbf{The effectiveness of PIE.} We test PIE on the re-ID benchmarks, and present the results in Table \ref{table:various_approaches} and Fig. \ref{fig:comparisons}.

Comparing with baseline 1 and baseline 2, we observe clearly that PIE yields higher re-ID accuracy. On Market-1501, for example, when using AlexNet and the FC7 descriptor, our method exceeds the two baselines by +5.5\% and +8,8\% in rank-1 accuracy, respectively. With ResNet-50, the improvement becomes slightly smaller, but still arrives at +5.0\% and +6.8\%, respectively. Specifically, rank-1 accuracy and mAP on Market-1501 arrive at 78.65\% and 53.87\%, respectively. On CUHK03 and VIPeR, consistent improvement over the baselines can also be observed.

Moreover, Figure \ref{fig:comparisons} shows that Kissme \cite{kostinger2012large} marginally improves the accuracy, proving that the PIE descriptor is well-learned. The concatenation of the Pool5 features of baseline 1 and 2 coupled with Kissme produces lower accuracy compared with ``PIE(Pool5)+kissme'', illustrating that the PBF network learns more effective embeddings than learning separately. We also find that the 2,048-dim ``PIE(Pool5,img)+EU'' and ``PIE(Pool5,pb)+EU'' outperforms the corresponding baseline 1 and 2. This suggests that PBF improves the baseline performance probably through the back propagation of the fused loss.

\begin{figure}
  \centering
  \includegraphics[width=2.9in]{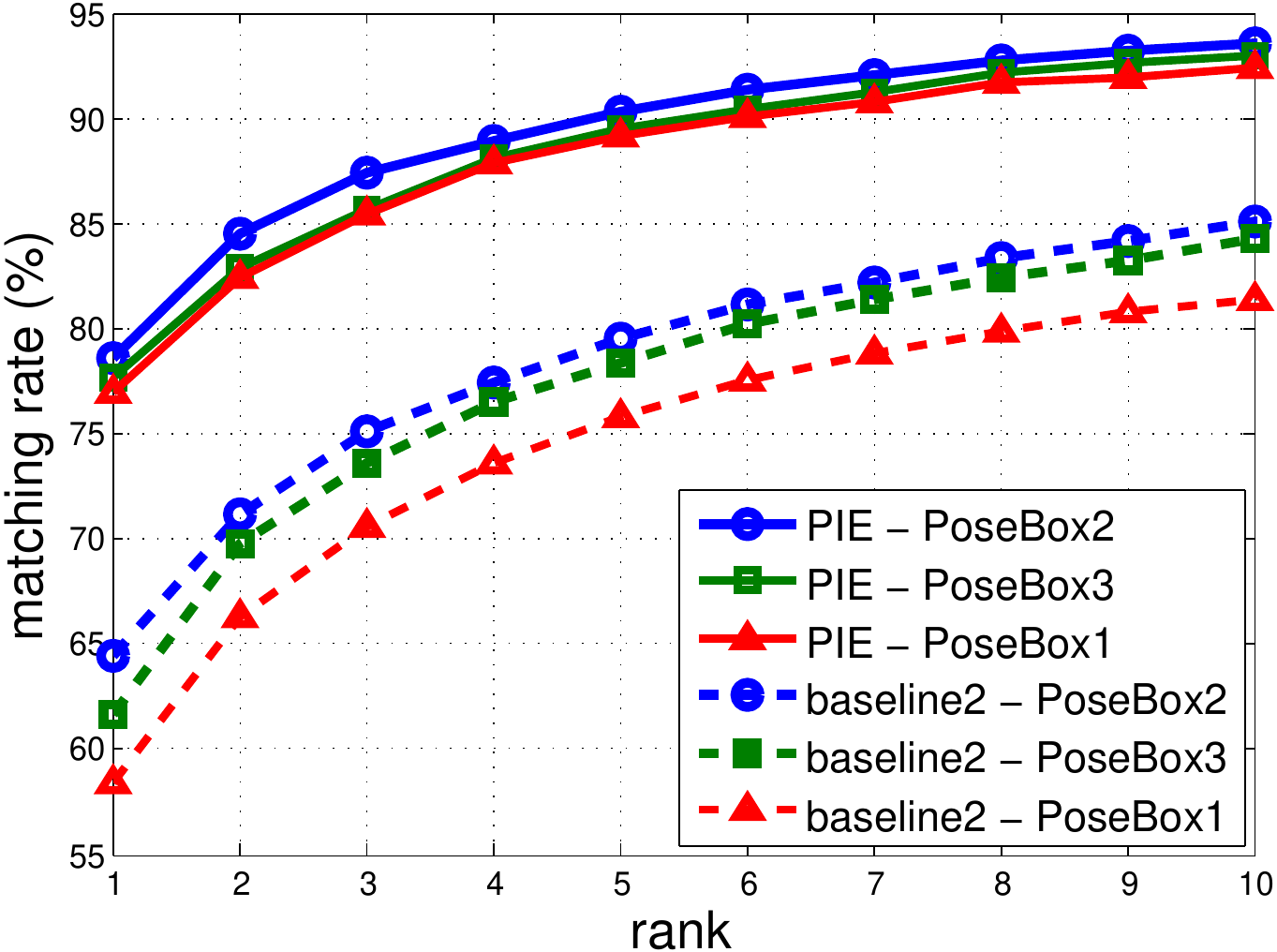}\\
  \caption{Re-ID accuracy of the three types of PoseBoxes. Results of both the baseline and PIE are presented on the Market-1501 dataset. }\label{fig:three_posebox}
\end{figure}

\textbf{Comparison of the three types of PoseBoxes.}
In Section \ref{sec:posebox}, three types of PoseBoxes are defined.
Their comparison results on Market-1501 are shown in Fig. \ref{fig:three_posebox}. Our observation is two-fold.

First, PoseBox2 is superior to PoseBox1. On Market-1501 dataset, PoseBox2 improves the rank-1 accuracy by xx\% over PoseBox1. The inclusion of arms therefore increases the discriminative ability of the system. Since the upper arm typically shares the same color/texture with the torso, we speculate that it is the long/short sleeves that enhance the descriptors. Second, PoseBox2 has better performance than PoseBox3 as well. For PoseBox3, the integration of the head introduces more noise due to the unstable head detection, which deteriorates the overall system performance. Nevertheless, we find in Fig. \ref{fig:three_posebox} that the gap between different PoseBoxes decreases after being integrated in PBF. It is because the combination with the original image reduces the impact of estimation errors and the information loss, a contribution mentioned in Section \ref{sec:intro}.

\begin{figure}
  \centering
  \includegraphics[width=2.9in]{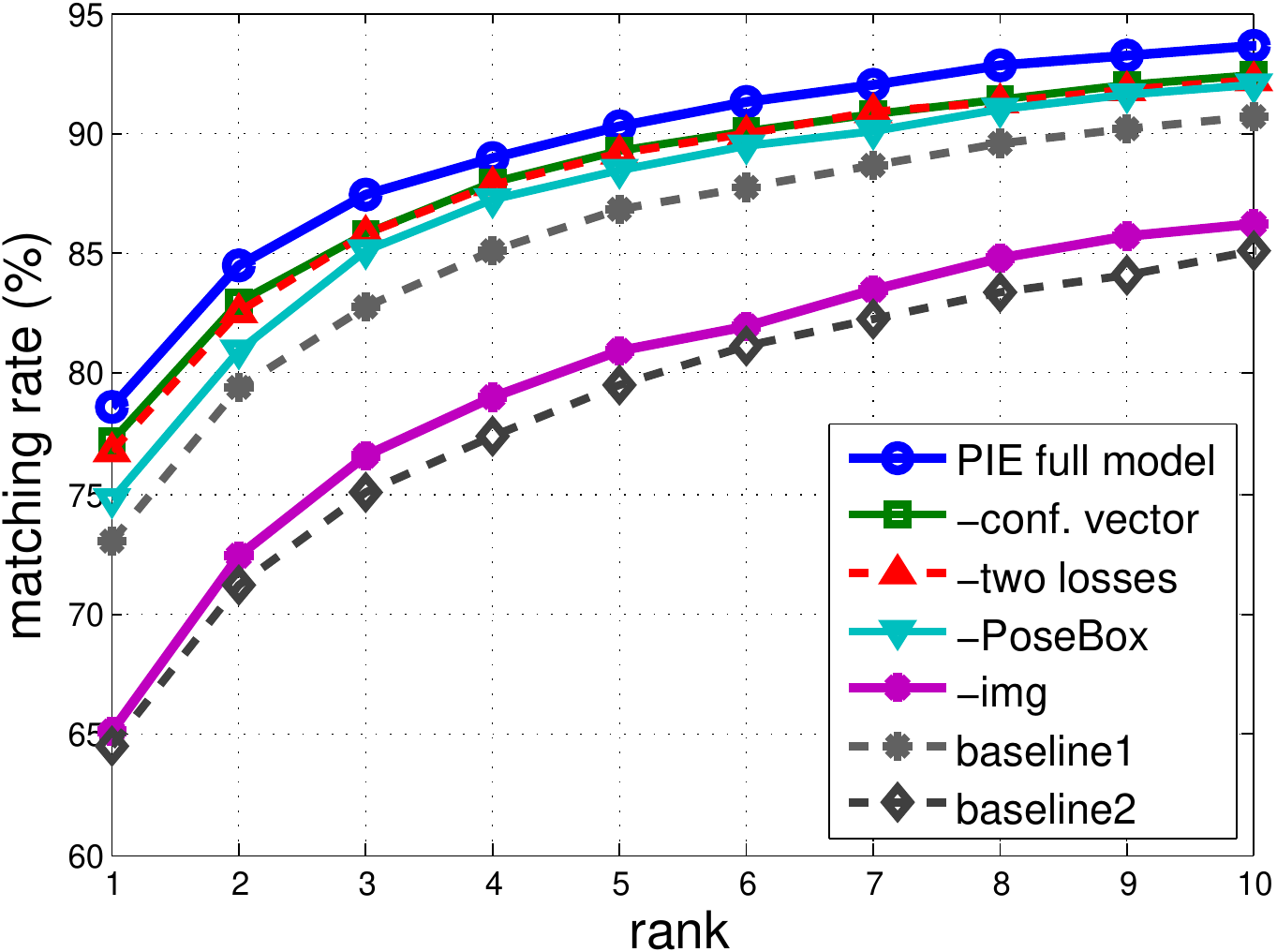}\\
  \caption{Ablation studies on Market-1501. From the ``full'' model, we remove one component at a time. The removed components include PoseBox, original image, the confidence vector, and the two losses of the PoseBox and the original image. }\label{fig:ablation}
\end{figure}

\textbf{Ablation experiment.}
To evaluate the effectiveness of different components of PBF, ablation experiments are conducted on the Market-1501 dataset. We remove one component from the full system at a time, including the PoseBox, the original image, the confidence vector, and the two losses of the PoseBox and original image streams.  The CMC curves are drawn in Fig. \ref{fig:ablation}, from which three conclusions can be drawn.

First, when the confidence vector or the two losses are removed, the remaining system is inferior to the full model, but displays similar accuracy. The performance drop is approximately 1\% in the rank-1 accuracy. It illustrates that these two components are important regularization terms. The confidence vector informs the system of the reliability of the PoseBox, thus facilitating the learning process. The two identification losses provide additional supervision to prevent the performance degradation of the two individual streams. Second, after the removal of the stream of the original image (``-img''), the performance drops significantly but still remains superior to baseline 2. Therefore, the original image stream is very important, as it reduces re-ID failures that likely result from pose estimation errors. Third, when the PoseBox stream is cut off (``-PoseBox''), the network is inferior to the full model, but is better than baseline 1. This validates the indispensability of PoseBox, and suggests that the confidence vector improves baseline 1.

\textbf{Comparison with the state-of-the-art methods.}
On Market-1501, we compare PIE with the state-of-the-art methods in Table \ref{table:sota_market}. It is clear that our method outperforms these latest results by a large margin. Specifically, we achieve \textbf{rank-1 accuracy = 77.97\%, mAP = 52.76\%} using the single query mode. To our knowledge, we have set new state of the art on the Market-1501 dataset.

On CUHK03, comparisons are presented in Table \ref{table:sota_CUHK03}. When metric learning is not used, our results are competitive in rank-1 accuracy with recent methods such as \cite{varior2016gated}, but are superior in rank-5, 10, 20, and mAP. When Kissme \cite{kostinger2012large} is employed, we report higher results: \textbf{rank-1 = 67.10\%, and mAP = 71.32\%}, which exceed the current state of the art. We note that in \cite{jose2016scalable}, very high results are reported on the hand-drawn subset but no results can be found on the detected set. We also note that metric learning yields smaller improvements on Market-1501 than CUHK03, because the PBF network is better trained on Market-1501 due to its richer annotations.

On VIPeR, we extract features using the off-the-shelf PIE model trained on Market-1501, and the comparison is shown in Table \ref{table:sota_viper}. We first compare PIE (using Euclidean distance) with the latest unsupervised methods, \eg, the Gaussian of Gaussian (GoG) \cite{matsukawa2016hierarchical}, the Bag-of-Words (BOW) \cite{zheng2015scalable} descriptors, \etc. We use the available code provided by the authors. We observe that PIE exceeds the competing methods in the rank-1, 5, and 10 accuracies. Then, compared with supervised works without feature fusion, our method (coupled with Mirror Representation \cite{chen2015mirror} and MFA \cite{yan2007graph}) has decent results. We further fuse the PIE descriptor with the pre-computed transferred deep descriptors \cite{wu2016enhanced} and the LOMO descriptor \cite{liao2015person}. We employ the mirror representation \cite{chen2015mirror} and the MFA distance metric coupled with the Chi Square kernel. The fused system achieves new state of the art on the VIPeR dataset with \textbf{rank-1 accuracy = 54.49\%}.

\setlength{\tabcolsep}{5pt}
\begin{table}[!t]
\footnotesize
\renewcommand{\arraystretch}{1.0}
\begin{center}
\caption{Comparison with state of the art on Market-1501.}
\begin{tabular}{l|ccccc}
\hline
\multirow{1}{*}{Methods} &
 rank-1 & rank-5 &rank-10& rank-20 &mAP\\
\hline
 BoW+Kissme \cite{zheng2015scalable} & 44.42 & 63.90&72.18 & 78.95&20.76\\

 WARCA \cite{jose2016scalable} &45.16&68.12&76&84&-\\
 Temp. Adapt. \cite{martinel2016temporal}& 47.92 & - &-&-&22.31\\
 SCSP \cite{chen2016similarity}& 51.90 & - &-&-&26.35 \\
 Null Space \cite{zhang2016learning}& 55.43& - & -&-&29.87\\
 LSTM Siamese \cite{varior2016siamese} &61.6&-&-&-&35.3\\
 Gated Siamese \cite{varior2016gated} & 65.88 &-&-&-&39.55\\
\hline
\hline
 \textbf{PIE (Alex)}& 65.68 & 82.51&87.89&91.63 &41.12\\
 \textbf{PIE (Res50)} & \textit{78.65} & \textit{90.26}&\textit{93.59} &\textit{95.69}& \textit{53.87}\\
  \textbf{+ Kissme} & \textbf{79.33} &\textbf{90.76}  & \textbf{94.41}&\textbf{96.52}&\textbf{55.95}\\
\hline
\end{tabular}
\label{table:sota_market}
\end{center}
\end{table}

\setlength{\tabcolsep}{4.5pt}
\begin{table}[!t]
\footnotesize
\renewcommand{\arraystretch}{1.0}
\begin{center}
\caption{Comparison with state of the art on CUHK03 (detected).}
\begin{tabular}{l|c|c|c|c|c}
\hline
\multirow{1}{*}{Methods} &
 rank-1 & rank-5 & rank-10& rank-20 &mAP\\
\hline
 BoW+HS \cite{zheng2015scalable} & 24.30 & - & -&-&-\\
 Improved CNN \cite{ahmed2015improved}&44.96&76.01&83.47&93.15&-\\
 XQDA  \cite{liao2015person}&46.25&78.90&88.55&94.25&-\\
 SI-CI \cite{wang2016joint}&52.2&74.3&92.3&-&-\\
 Null Space \cite{zhang2016learning}& 54.70& 84.75 &94.80&95.20 &-\\
 LSTM Siamese \cite{varior2016siamese} &57.3&80.1&88.3&-&46.3\\
 MLAPG \cite{liao2015efficient} &57.96 & 87.09 &94.74&98.00&-\\
 Gated Siamese \cite{varior2016gated} & 61.8 &80.9&88.3&-&51.25\\
\hline
\hline
 \textbf{PIE (Alex)} & 62.60 & 87.05&92.50 &96.30&67.91\\
 \textbf{PIE (Res50)} &\textit{61.50} &\textit{89.30} &\textit{94.50}&\textit{97.60}&\textit{67.21} \\
  \textbf{+ Kissme} & \textbf{67.10} & \textbf{92.20} & \textbf{96.60} & \textbf{98.10} & \textbf{71.32}\\
\hline
\end{tabular}
\label{table:sota_CUHK03}
\end{center}
\end{table}

\setlength{\tabcolsep}{4.9pt}
\begin{table}[!t]
\footnotesize
\renewcommand{\arraystretch}{1.0}
\begin{center}
\caption{Comparison with state of the art on VIPeR. The top 6 rows are unsupervised; the bottom 10 rows use supervision. }
\begin{tabular}{l|c|c|c|c}
\hline
\multirow{1}{*}{Methods} &
 rank-1 & rank-5 & rank-10& rank-20 \\
\hline
 GOG \cite{matsukawa2016hierarchical}&21.14  & 40.34 & 53.29&\textbf{67.21} \\
 Enhanced Deep \cite{wu2016enhanced}& 15.47 & 34.53 & 43.99& 55.41\\
 SDALF \cite{ma2012bicov}& 19.87 & 38.89 & 49.37& {65.73}\\
 gBiCov \cite{ma2012bicov}& 17.01 & 33.67 & 46.84& 58.72\\
 BOW \cite{zheng2015scalable}& 21.74 & - & -& 60.85\\
 \hline
 \textbf{PIE} & \textbf{27.44} & \textbf{43.01} & \textbf{50.82}&{60.22} \\
\hline
\hline
 XQDA  \cite{liao2015person}&40.00&67.40&80.51&91.08\\
 MLAPG \cite{liao2015efficient} &40.73 & - &82.34&92.37\\
 WARCA \cite{jose2016scalable} &40.22 & 68.16& 80.70 &91.14\\
 Null Space \cite{zhang2016learning}& 42.28& 71.46 & 82.94&92.06 \\
 SI-CI \cite{wang2016joint}&35.8&67.4&83.5&-\\
 SCSP \cite{chen2016similarity} & 53.54 &82.59&91.49& 96.65\\
 Mirror \cite{chen2015mirror} & 42.97 & {75.82} &87.28& 94.84\\
 Enhanced \cite{wu2016enhanced} + Mirror \cite{chen2015mirror} & 34.87 & 66.68 &79.30 & 90.38\\
 LSTM Siamese \cite{varior2016siamese} &42.4&68.7&79.4&-\\
 Gated Siamese \cite{varior2016gated} & 37.8 &66.9&77.4&-\\

\hline
\textbf{PIE+Mirror \cite{chen2015mirror}+MFA\cite{yan2007graph} }& 43.29 & 69.40 & 80.41&89.94\\
\textbf{Fusion+MFA }& \textbf{54.49} & \textbf{84.43} & \textbf{92.18}&\textbf{96.87}\\
\hline
\end{tabular}
\label{table:sota_viper}
\end{center}
\end{table}

\begin{figure}
  \centering
  \includegraphics[width=3.3in]{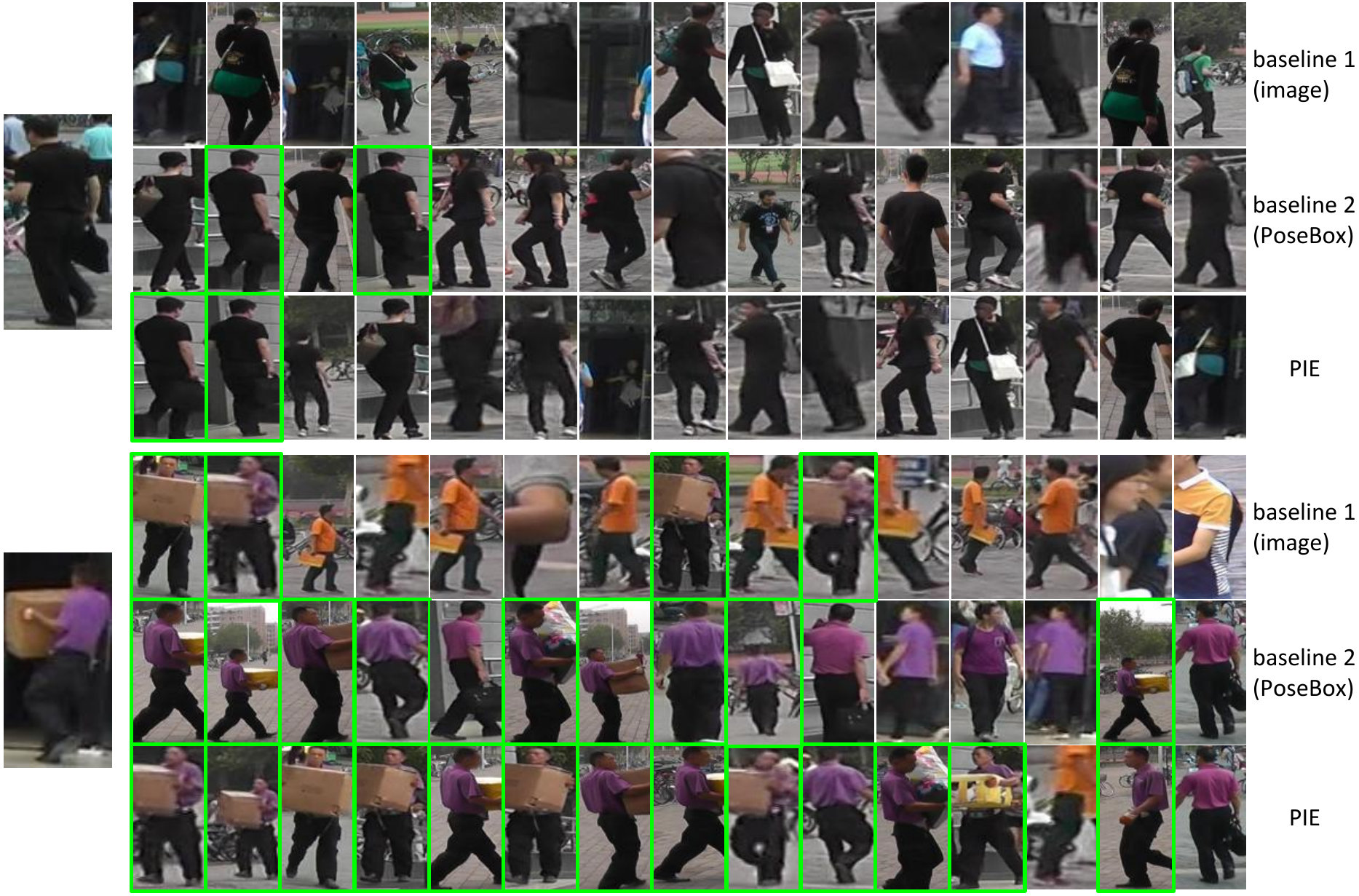}\\
  \caption{Sample re-ID results on the Market-1501 dataset. For each query placed on the left, the three rows correspond to the rank lists of baseline 1, baseline 2, and PIE, respectively. Green bounding boxes denote correctly retrieved images. }\label{fig:sample_results}
\end{figure}

Two groups of  sample re-ID results are shown in Fig. \ref{fig:sample_results}. In the first query, for example, the cyan clothes on the background lead to the misjudgement of the foreground characteristics, so that some pedestrians with local green/blue colors incorrectly receive top ranks. Using PIE, foreground can be effectively cropped, leading to more accurate pedestrian matching.

\section{Conclusion}\label{sec:conclusion}
This paper explicitly addresses the pedestrian misalignment problem in person re-identification. We propose the pose invariant embedding (PIE) as pedestrian descriptor. We first construct PoseBox with the 16 joints detected with the convolutional pose machine \cite{wei2016convolutional}. PoseBox helps correct the pose variations caused by camera views, person motions and detector errors and enables well-aligned pedestrian matching. PIE is thus learned through the PoseBox fusion (PBF) network, in which the original image is fused with the PoseBox and the pose estimation confidence. PBF reduces the impact of pose estimation errors and detail loss during PoseBox construction. We show that PoseBox yields fair accuracy when used alone and that PIE produces competitive accuracy compared with the state of the art.


{\small
\bibliographystyle{ieee}
\bibliography{egbib}
}

\end{document}